\documentclass[10pt,twocolumn,letterpaper]{article}

\makeatletter
\@namedef{ver@everyshi.sty}{}
\makeatother

\usepackage{abstract}
\usepackage{cvpr}              

\usepackage{graphicx}
\usepackage{color}
\usepackage{amsmath,cases}
\usepackage{amssymb}
\usepackage{booktabs}

\usepackage{color}

\usepackage{comment}
\usepackage{subfig}
\usepackage{wrapfig}
\usepackage{nicefrac}
\usepackage{float}
\usepackage{animate}
\usepackage[export]{adjustbox}
\usepackage[normalem]{ulem}
\usepackage{multirow}
\usepackage{bbold}
\usepackage[font=small]{caption}
\usepackage{bbold}
\usepackage{enumitem}

\usepackage{tikz}
\usetikzlibrary{calc}
\usetikzlibrary{positioning}
\usepgflibrary{arrows}

\usepackage[pagebackref,breaklinks,colorlinks]{hyperref}

\usepackage[capitalize]{cleveref}
\crefname{section}{Sec.}{Secs.}
\Crefname{section}{Section}{Sections}
\Crefname{table}{Table}{Tables}
\crefname{table}{Tab.}{Tabs.}

\def\cvprPaperID{955} 
\def\confName{CVPR}
\def\confYear{2022}

\begin{document}

\title{Watch It Move: Unsupervised Discovery of 3D Joints for Re-Posing of Articulated Objects}

\author{
  Atsuhiro Noguchi\textsuperscript{\dag, \ddag}\thanks{Work partially done when Atsuhiro Noguchi was an intern at NVIDIA.}\hspace{4mm}Umar Iqbal\textsuperscript{\dag}\hspace{4mm}Jonathan Tremblay\textsuperscript{\dag}\hspace{4mm}Tatsuya Harada\textsuperscript{\ddag,\S}\hspace{4mm}Orazio Gallo\textsuperscript{\dag}  \\
  \textsuperscript{\dag}NVIDIA\hspace{5mm}   \textsuperscript{\ddag}The University of Tokyo\hspace{5mm}   \textsuperscript{\S}RIKEN
}


\def\traj{\mathcal{T}_\Theta} 
\def\decoder{\mathcal{S}_\Theta} 
\def\ellipsoids{\mathcal{E}} 
\def\object{\mathcal{O}} 
\def\rotmat{{\bf R}}
\def\partcenter{{\bf t}}
\def\joint{{\boldsymbol \xi}}
\def\jointStruct{\Gamma}
\def\pointcolor{{\bf c}}

\twocolumn[{
      \vspace{-7mm}
      \renewcommand\twocolumn[1][]{#1}%
      \maketitle
      \begin{center}
        \centering
        \vspace*{-9mm}
        \captionsetup{type=figure}

\begin{tikzpicture}[inner sep=0pt,line width=0.5pt]

    \newlength{\inputHeight}
    \setlength{\inputHeight}{35pt}
    \newlength{\singleCropWidth}
    \setlength{\singleCropWidth}{80pt}

    \newlength{\inputOffset}
    \setlength{\inputOffset}{15pt}

    \newlength{\labelHeight}
    \setlength{\labelHeight}{0.5in}

    \newlength{\hgap}
    \setlength{\hgap}{5pt}

    \newlength{\arrowLength}
    \setlength{\arrowLength}{20pt}

    \setlength{\fboxsep}{0pt}
    \setlength{\fboxrule}{0.25pt}

    \node(leftLimit) at (0,0){};

    \node(rightLimit) at ($(leftLimit)+(\textwidth,0)$){};

    \node(tZero)[anchor=west] at (leftLimit){\fbox{\includegraphics[height=\inputHeight]{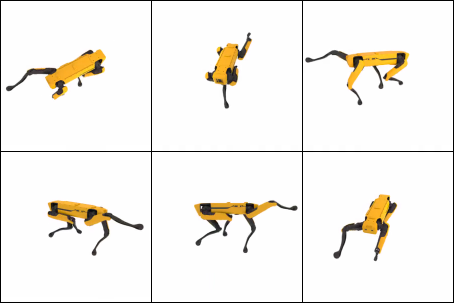}}};
    \node(tOne) [below right = \inputOffset and \inputOffset of tZero, anchor=south east]{\fbox{\includegraphics[height=\inputHeight]{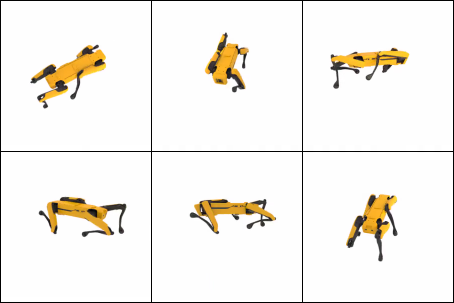}}};
    \node(tT) [below right = \inputOffset and \inputOffset of tOne, anchor=south east]{\fbox{\includegraphics[height=\inputHeight]{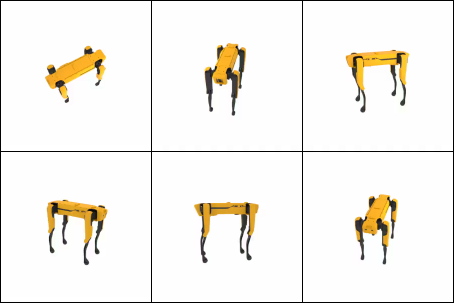}}};
    \node(inputLabel) [below = 22pt of tOne]{(a) Multi-View Video\strut};
    \path let \p1 = (tOne), \p2 = (tT.east) in node(a) at ($(\x2,\y1) + (\hgap,0)$){};
    \path let \p1 = (tOne), \p2 = (tT.east) in node(b) at ($(\x2,\y1) + (\hgap+\arrowLength,0)$){};
    \draw[round cap-latex](a)--(b);

    \node(reconstruction_back)[right= \hgap of b]{\includegraphics[width=1.8\singleCropWidth]{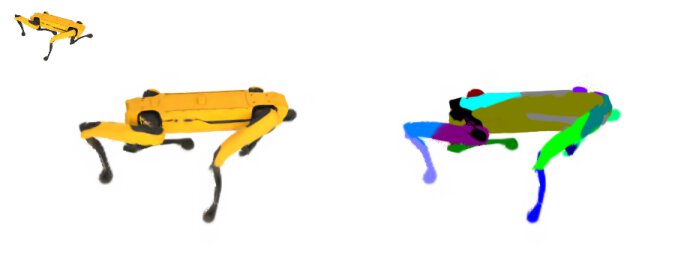}};
    \node(reconstruction)[right= \hgap of b]{\animategraphics[width=1.8\singleCropWidth,poster=00007]{10}{figures/teaser/spot/pane_topleft_lowrez_jpg/}{00000}{00058}};

    \node(reconsLabel) at (inputLabel.east -| reconstruction.south) {(b) RGB and Parts Reconstruction\strut};
    \node(gtLabel) at ($(reconstruction.north west)+(-5pt,5pt)$) [anchor = west] {\tiny Original Frame\strut};

    \node(c)[right = \hgap of reconstruction, anchor=west] {};
    \node(d) at ($(c)+(\arrowLength,0)$){};
    \draw[round cap-latex](c)--(d);

    \node(injectTo) at ($(c)!0.5!(d)$){};
    \node(injectFrom) at ($(injectTo)-(0,10pt)$){};
    \node(pose)[below= 0pt of injectFrom]{\tiny{\begin{tabular}{c} New Pose\\ $\bigg\{\rotmat_i, \partcenter_i \bigg\}$ \end{tabular}}};
    \draw[round cap-latex](injectFrom)--(injectTo);

    \node(repose_back)[right= \hgap of d]{\includegraphics[width=\singleCropWidth]{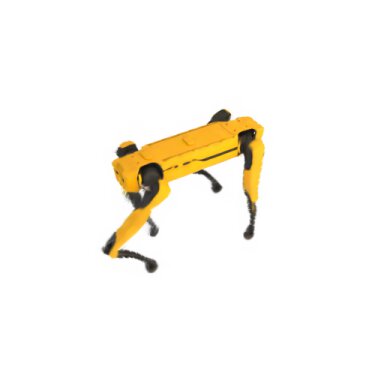}};
    \node(repose)[right= \hgap of d]{\animategraphics[width=\singleCropWidth, poster=00013]{10}{figures/reposing_results/spot_rotate_lowrez_jpg/gen_}{00000}{00079}};
    \node(reposeLabel)  at (reconsLabel.east -| repose.south) {(c) Re-Posing\strut};

    \newlength{\horizSepSpace}
    \setlength{\horizSepSpace}{2pt}

    \node(vertTop) [right = \horizSepSpace of repose.north east]{};
    \node(vertBot) [right = \horizSepSpace of repose.south east]{};
    \draw[round cap-round cap](vertTop)--(vertBot);

    \node(reposeRobotGT)[right = \horizSepSpace of vertTop, anchor=north west]{\includegraphics[width=0.5\singleCropWidth]{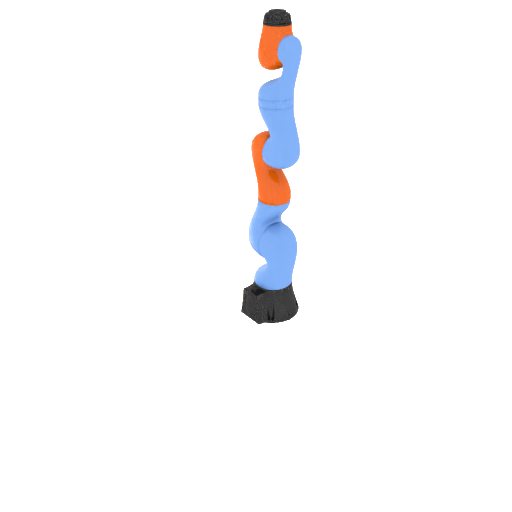}};
    \node(reposePersonGT)[right = \horizSepSpace of vertBot, anchor=south west]{\includegraphics[width=0.5\singleCropWidth]{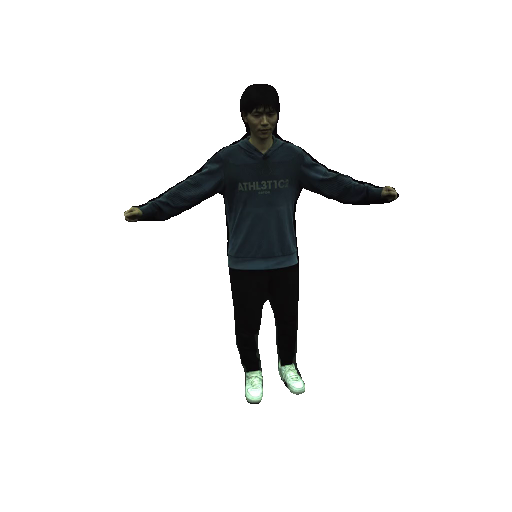}};
    \node(rightmostPane_back) at (reposeRobotGT.north east) [anchor=north west]{\includegraphics[width=\singleCropWidth]{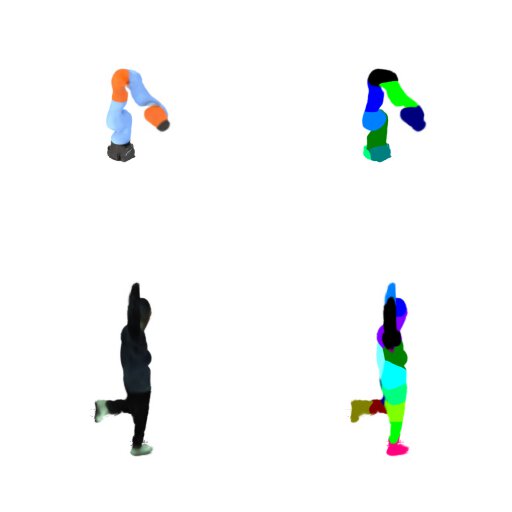}};
    \node(rightmostPane) at (reposeRobotGT.north east) [anchor=north west]{\animategraphics[width=\singleCropWidth, poster=00020]{10}{figures/teaser/rightmost_pane_lowrez_jpg/}{00000}{00079}};
    \node(tmpNode) at ($(rightmostPane.south west)!0.25!(rightmostPane.south east)$){};
    \node(rightmostLabel)  at (reposeLabel.east -| tmpNode) {(d) Other Categories\strut};


\end{tikzpicture}

        \vspace{-3mm}
        \captionof{figure}{\textbf{Animated figure (view in Adobe Reader and click on panes (b), (c), and (d)).} Our method learns to render novel views of an articulated, moving object by ``watching'' it move in a multi-view video sequence with associated foreground masks, as shown in the animation in (b).
          Simultaneously, it discovers the object's parts and joints with \emph{no additional supervision}. The learned structure can be used to \emph{explicitly} re-pose the object, by roto-translating each part around its joint. In panes (c) and (d) we re-pose objects from multiple categories to configurations never seen in training, an operation only possible thanks to the structure we discover from the input videos.
          \label{fig:teaser}}
      \end{center}
    }]

\maketitle

\begin{abstract}

Rendering articulated objects while controlling their poses is critical to applications such as virtual reality or animation for movies.
Manipulating the pose of an object, however, requires the understanding of its underlying structure, that is, its joints and how they interact with each other.
Unfortunately, assuming the structure to be known, as existing methods do, precludes the ability to work on new object categories.
We propose to learn both the appearance and the structure of previously unseen articulated objects by observing them move from multiple views, with no joints annotation supervision, or information about the structure.
We observe that 3D points that are static relative to one another should belong to the same part, and that adjacent parts that move relative to each other must be connected by a joint.
To leverage this insight, we model the object parts in 3D as ellipsoids, which allows us to identify joints.
We combine this explicit representation with an implicit one that compensates for the approximation introduced.
We show that our method works for different structures, from quadrupeds, to single-arm robots, to humans.
\noindent{The code is available at \url{https://github.com/NVlabs/watch-it-move}}

\end{abstract}
\vspace{-6mm}

\saythanks

\section{Introduction}\label{sec:intro}

Using images to infer both the appearance and the functional structure of generic, real-world objects is a fundamental goal of computer vision.
From a practical standpoint, it would allow us to render and manipulate physical objects in the \emph{metaverse}.
But its appeal goes further, as it requires pushing the boundaries of our ability to learn from data with no direct supervision.

Our community made dramatic progress towards appearance capture and novel view synthesis, particularly for static scenes~\cite{mildenhall2020nerf,ruckert2021adop,lin2021barf,choi2019extreme,yu2021plenoctrees,barron2021mip,zhou2018stereo}.
Several recent methods can also capture dynamic scenes and \emph{reenact} their motion~\cite{pumarola2021DNERF,flow-fields,park2021hypernerf,wang2021dctnerf,nr-nerf}.
We use the term ``reenacting'' to highlight that these methods cannot explicitly control the pose of the dynamic objects.
Rather, they replay through the poses that were observed.
\emph{Re-posing} an articulated object---\ie, the explicit manipulation of its pose---requires knowing the location of the joints and how the different parts of the object interact with each other\footnote{Image-to-image translation methods (e.g.,\cite{ma2017pose}) can also re-pose, but we focus on methods that allow for the explicit definition of the target pose.
}.
Learning to predict the location of joints in 3D is a well-studied task, at least for humans, and it is generally tackled using 2D~\cite{rohdin2018multiview, yao2019monet, iskakov2019learnable, kocabas2019epipolar, iqbal2020learning, wandt2021canonpose} or 3D~\cite{sun18integeral, iqbal2018hand, Zimmermann2017ICCV, labbe2021robopose, lee2020icra:dream, hmrKanazawa18, li2020hybrik} ground truth information.
When not using joints supervision, existing pose manipulation methods rely on a predefined model, that is, a template structure~\cite{schmidtke2021unsupervised,kundu2020appearance}.
However, annotations are expensive and object-specific, which is why they are only available for limited classes of objects, such as people or faces~\cite{sanyal2019learning,peng2021neural,h36m_pami}.

We aim at re-posing an articulated object from a category not seen before, using only a multi-view video and corresponding foreground mask, as shown in Figure~\ref{fig:teaser}.
Our approach requires no additional supervision, no prior knowledge about the structure, nor networks pre-trained on auxiliary tasks: we learn the appearance and the structure of the object by just watching it move.
Like existing methods~\cite{deng2020nasa,noguchi2021neural}, to express explicit pose changes, we treat the articulated object as a set of posed parts, each connected to other parts through joints.
However, rather than relying on direct supervision, we note that a joint is a 3D point around which a part must rotate to produce the piece-wise, rigid deformation observed in the input images.
This allows us to get indirect supervision for the locations of the joints from the image reconstruction loss.

Our approach, inspired by neural implicit representations, is scene-specific and predicts the color and the signed-distance function (SDF) of any 3D point, allowing us to generate any desired frame by volumetric rendering~\cite{wang2021neus}.
We also learn certain properties of the object \emph{explicitly}.
Specifically, we model the object as a set of ellipsoids.
A functional part of the object can be represented by one or more ellipsoids, as shown in Figure~\ref{fig:parametrization}.
We optimize the geometric properties of the ellipsoids, \ie, their size and pose, for each frame of the input sequence.
The color and density of a 3D point, then, can be predicted from the combined contribution of the ellipsoids.
Because these ellipsoids only afford a coarse approximation of the object, we also estimate a residual with respect to this explicit part-based representation.
In addition to regularizing the optimization landscape, this representation provides a key advantage:
the relative motion of the parts can be explicitly observed over time, which offers clues on the locations of the joints.
Note that this applies to unobserved categories, and requires no prior knowledge on the number of parts that compose it.
Because we do not use any prior on the structure of the object or supervision annotations, our method can re-pose any articulated object from a single multi-view video sequence and the corresponding foreground masks.
The pose of the object can be manipulated by applying the appropriate roto-translation to the different joints.
Figures~\ref{fig:teaser}(c) and (d) show examples of object re-posing for different categories, structures, and number of parts---all of which were unknown at training time.
Our method
\begin{itemize}[noitemsep,topsep=0pt,parsep=0pt,partopsep=0pt,leftmargin=*]
  \item Is the first to learn a re-poseable shape representation from multi-view videos and foreground masks, without additional supervision or prior knowledge of the underlying structure,
  \item it discovers the number and location of physically meaningful joints---also learned with no annotations, and
  \item it is structure agnostic and can thus be learned for previously unseen articulated object categories.
  \item Our reconstruction and re-posing results are on par or better than those of category-specific methods that use prior knowledge.
\end{itemize}

\section{Related Work}\label{sec:related}

\subsection{Object Re-Posing and Novel-View synthesis}
Synthesizing images of articulated objects under novel poses and viewpoints
is critical to several applications. Earlier methods formulated
the problem as conditional image-to-image
translation~\cite{ma2017pose,yang2020towards,zhu2019progressive,esser2018variational,pumarola2018unsupervised,chan2019dance,lwb2019,yoon2021poseguided}.
Given an image of an object and a target pose, these methods use a generator
model to transfer a given image to a target pose.
The conditioning pose is usually obtained from 2D keypoints or parametric meshes.
However, keypoints or mesh models are available for handful of object categories
(\eg, faces, human body, and hands), preventing these methods from generalizing to
arbitrary object classes.

More recently, NeRF~\cite{mildenhall2020nerf} ignited a wave of research
on synthesizing novel views of an object by using a sparse set of
multi-view images
~\cite{mildenhall2020nerf,lin2021barf,yu2021plenoctrees,barron2021mip,oechsle2021unisurf, wang2021neus}.
These methods learn an implicit 3D representation that provides
the color and density of each point in 3D space. Photorealistic images can
then be generated using volumetric rendering. Since the implicit 3D representation
they use is continuos and topology-agnostic, these methods can reconstruct
arbitrary, \emph{static} objects. 
Many follow-up works extend NeRF to model dynamic scenes, using single- or multi-view videos for training~\cite{pumarola2021DNERF,
    flow-fields,park2021hypernerf,wang2021dctnerf,nr-nerf}.
However, these methods only ``reenact'' the video used for training, and do not offer
control over the articulated pose. We build on these developments
and propose a method that also provides control over the articulated pose of the objects.

\begin{figure}
    \centering
    \includegraphics[width=\columnwidth]{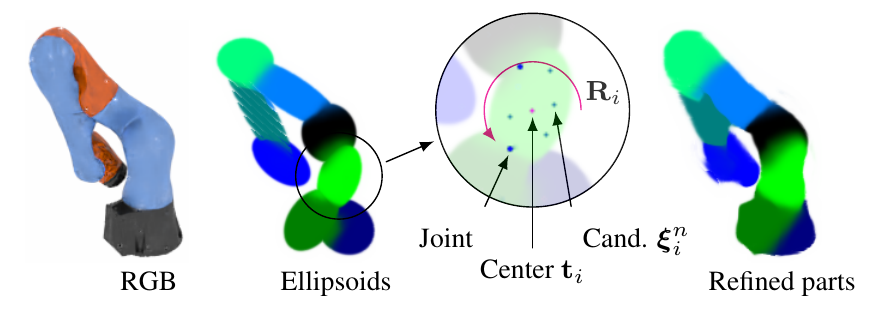}
    \caption{We explicitly represent each object's part as an ellipsoid centered at $\partcenter_i$ and oriented with $\rotmat_i$ (magenta arrow). We identify the part's joints from a pool of candidates, $\joint_i^n$. The final reconstruction is obtained by estimating a residual \wrt the ellipsoids.}
    \label{fig:parametrization}
    \vspace{-15pt}
\end{figure}

\begin{figure*}
    \centering
    \includegraphics[width=\textwidth]{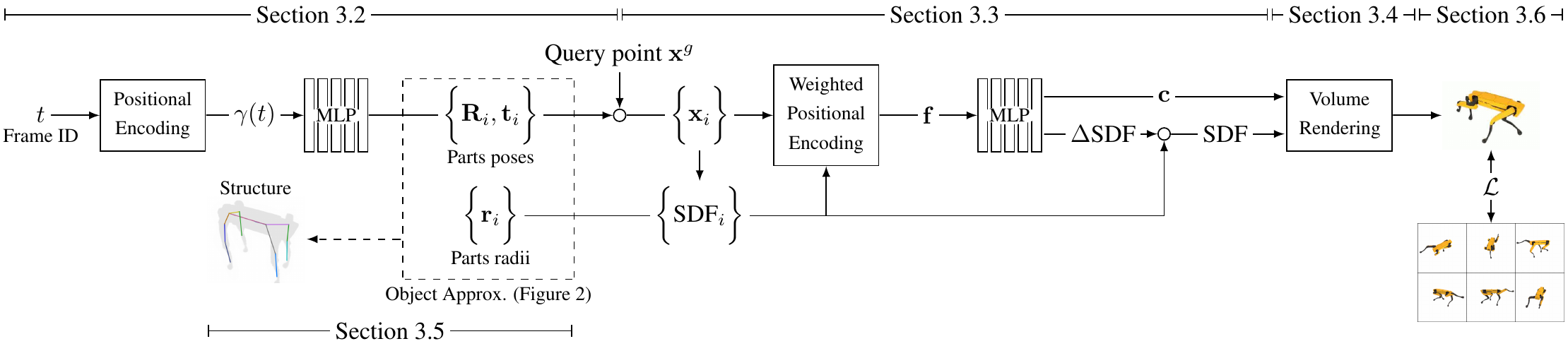}
    \caption{Given a multi-view video of an articulated, dynamic object, an MLP associates a frame id with the configuration of the object's parts, approximated with ellipsoids. To predict the color of a 3D query point in global coordinates, we assemble a feature ${\bf f}$ from its coordinates and signed-distance functions, expressed with respect to all of the parts. From ${\bf f}$, a second MLP produces the color and a residual SDF to perform volume rendering. Simultaneously, from the parts we discover the structure. We train the system end to end, thus back-propagating the error all the way to the poses and radii of the parts. For clarity, some components and loss terms are not shown here.}
    \label{fig:architecture}
\end{figure*}

To allow re-posing the objects, various implicit
representations for articulated objects have also been proposed,
especially for humans. They allow novel view and pose synthesis,
but require ground truth poses~\cite{deng2020nasa, noguchi2021neural,su2021anerf},
or dense 3D meshes~\cite{tiwari2021neural, peng2021animatable, peng2021neural, kwon2021neural,
    liu2021neural}
annotations for the training image.
In contrast, we propose
a re-poseable 3D implicit representation trained only from multi-view videos and foreground masks of a previously unseen object category.
We simultaneously decompose the parts, estimate the connections between them,
and reconstruct the image 
with no prior information about the structure of the object.

\subsection{Discovery of 3D Joints of Articulated Objects}
Explicitly re-posing an object is straightforward if the joints locations are given, but localizing the 3D joints is challenging.
Ground-truth 3D joints supervision simplifies the problem~\cite{sun18integeral, iqbal2018hand, Zimmermann2017ICCV, labbe2021robopose, lee2020icra:dream, hmrKanazawa18, li2020hybrik}.
However, 3D annotations are expensive to gather and, perhaps more importantly, they make the resulting algorithms category-specific.
Other works simplify the problem and rely on 2D annotations and
multi-view~\cite{rohdin2018multiview, yao2019monet, iskakov2019learnable, kocabas2019epipolar, iqbal2020learning, wandt2021canonpose}
or temporal~\cite{novotny2019C3DPO} information for 3D supervision.
Although 2D joints are cheaper to annotate, the process is
still time-consuming and hard to scale to a large number of objects and classes.
To address this, some recent methods aim to discover the joints of
articulated objects using self-supervised learning~\cite{jakab2020self, kundu2020appearance,schmidtke2021unsupervised}.
While these methods show impressive results, they still rely on carefully
designed, object-specific templates and/or prior information, which cannot
be directly applied to other object classes.
Other methods can handle arbitrary
objects but provide only 2D landmarks~\cite{thewlis2017unsupervised, zhang2018unsupervised, lorenz2019unsupervised, choudhury2021unsupervised}.
There exist some works that discover 3D
keypoints using self-supervision and multi-view data, but they are limited
to rigid objects~\cite{suwajanakorn2018discovery}.
Other works estimate the parts and the structure of arbitrary objects from videos, but they require ground truth 2D trajectory of keypoints~\cite{yan2008factorization,fayad2011automated}.
In contrast to these methods, our approach discovers 3D joints of articulated objects
and does not require any 2D or 3D annotations, predefined template, or any other prior
knowledge about the object, which makes it category-agnostic.
Additionally, most of these methods
only provide locations for a sparse set of keypoints/joints and do not provide any
information about the surface geometry or texture of the object.
While some methods provide self-supervised dense part labels,
they are limited to 2D information~\cite{hung:CVPR:2019,sabour2021flowcaps}.
In contrast, our re-posable shape representation
provides dense part labels, 3D surface geometry, as well as the texture of
each part (implicitly).

Our work is also related to the recent methods for 3D shape representation that use
implicit functions~\cite{genova2019learning, genova2020local, bozic2021neural},
which represent a deformable 3D shape as a composition of simple shape
elements, \eg, 3D Gaussians (where the level set is an ellipsoid). Each element
contributes to the implicit surface of the shape.
However, these methods
cannot learn surface textures, they require the ground truth 3D shape for training,
and do not learn the physical connectivity
between parts, which prevents explicit re-posing.
Concurrent methods circumvent the need of 3D shape supervision, but do not allow for explicit re-posing~\cite{yang2021viser,yang2021banmo}.

\section{Method}\label{sec:overview}

\subsection{Overview}\label{sec:method_overview}
Our method takes as input a sequence of $T$ multi-view, posed images of an articulated, moving object, $\object$, and foreground masks indicating its silhouette.
From those, we learn to render novel views of $\object$.
We also discover plausible joints that allow us to render $\object$ in a new pose and from a novel viewpoint, \emph{without any additional supervision}.
We use a hybrid representation of the object that combines an \emph{explicit} rough approximation of its body, and a subsequent \emph{implicit} refinement.
More concretely, we represent $\object$ explicitly as a set of $P$ parts, each approximated with an ellipsoid, see Figure~\ref{fig:parametrization}.
Rather than assuming $P$ to be known, we over-segment the object and subsequently merge the different parts as needed (Section~\ref{sec:merging}).
The ellipsoid representing part $i$ is parametrized with its three-dimensional radius ${\bf r}_i$.
(Throughout the paper, we use bold for vectors and matrices.)
Its pose at time $t$ is represented by the translation of its center of mass, $\partcenter_i(t)$, and a rotation matrix, $\rotmat_i(t)$.
To discover and localize the object's joints, which define the relationship between different parts, we observe that a 3D point is a meaningful joint if roto-translating a part around it explains a pose change in the reconstructed image.

There are four components to our method.
First, a trainable module estimates the pose of each part, at each frame $t\in[1,T]$ (Section~\ref{sec:trajectory}).
From the posed parts, we propose to discover the underlying structure (Section~\ref{sec:structure}), which we use as regularization during training, and to re-pose the object at inference.
The third is a module that also uses the pose of the parts, and is trained to predict the color and signed-distance function of points in 3D (Section~\ref{sec:decoder}).
The final component renders the output view by performing volumetric rendering on these predictions~\cite{mildenhall2020nerf,wang2021neus} (Section~\ref{sec:rendering}).
We train the system end-to-end (Figure~\ref{fig:architecture}).

\subsection{Pose Estimation}\label{sec:trajectory}
As shown in Figure~\ref{fig:parametrization}, we represent each part of the object $\object$ as an ellipsoid $e_i$, such that their union, $\ellipsoids$, approximates the object's 3D shape $\Omega$: $\ellipsoids = \bigcup\limits_{i=1}^{P} e_i \approx \Omega$.
Each ellipsoid has a learnable three-dimensional radius parameter ${\bf r}_i$.
Using the frame id $t$ as input, we train an MLP, $\traj$, that outputs
the global rotation $\rotmat_i(t)$ (represented with a $3\times3$ rotation matrix) and translation $\partcenter_i(t)$ of each $e_i$.
We initialize these parameters randomly and observe that the optimization is reasonably robust to different initializations.
Following common practice~\cite{mildenhall2020nerf}, rather than feeding $t$ to $\traj$ directly, we use positional encoding $\gamma(t)$, where  $\gamma(\cdot) = \{\cos(\alpha~\cdot)\}_{\{\alpha=1:50\}}$.
Since we overfit our system to a single scene, we can directly optimize the rotation and translation of each part in the global coordinate system, for each time frame: $\traj: \gamma(t) \rightarrow \{\rotmat_i(t), \partcenter_i(t)\}_{\{i=1:P\}}$.
By predicting rotations and translations in the global coordinate system, we naturally force the pose of the object to be estimated consistently across views.

\subsection{Shape and Appearance Decoder}\label{sec:decoder}
Similar to NeuS~\cite{wang2021neus}, we seek to estimate the color, $\pointcolor$, and signed-distance functions (SDFs), $d$, at any 3D point, ${\bf x}^g$, to perform volumetric rendering.
Since the ellipsoids alone cannot accurately capture the object's shape, we use a second MLP, $\decoder$, to predict a residual.
To ensure that the final shape does not deviate significantly from $\ellipsoids$, we represent this as a residual SDFs, $\Delta d$, which is bounded by construction.
We first convert the query point ${\bf x}^g$, expressed in global coordinates, to the local coordinate system of each part
${\bf x}_i(t) = (\rotmat_i(t))^{-1}({\bf x}^g - \partcenter_i(t))$,
and we apply weighted positional encoding~\cite{noguchi2021neural} to compute a feature vector
\begin{equation}\label{eq:positional}
    {\bf f} = \textsc{cat}\{\text{w}^\text{PE}_i ~ \gamma({\bf x}_i(t))\}_{\{i=1:P\}},
\end{equation}
where $\textsc{cat}$ is the concatenation operation.
The weights in Equation~\ref{eq:positional} are computed as
\begin{equation}\label{eq:weights}
    {\bf w}^{\text{PE}} = \text{softmax}\left\{-s^{\text{PE}} d_i\right\}_{\{i =1:P\}},
\end{equation}
where $d_i = \text{SDF}_i({\bf x}_i(t), e_i)$, and $s^{\text{PE}}$ is a learnable temperature parameter for the softmax.
The SDFs from the ellipsoids can be computed directly from their radii and poses (see the appendix).
Note that ${\bf f}$ effectively subsumes the current estimates of the ellipsoids, their pose, and the location of the sampled point.
We then feed ${\bf f}$ to a second MLP
\begin{equation}\label{eq:decoder_mlp}
    \decoder: {\bf f}({\bf x}^g) \rightarrow (\pointcolor, \widetilde{\Delta d})\big\rvert_{{\bf x}^g},
\end{equation}
where $\pointcolor$ is color of ${\bf x}^g$, and $\widetilde{\Delta d}$ a residual with respect to the SDFs estimated from the ellipsoids, which we compress as
\begin{equation}\label{eq:compress}
    \Delta d  = d_{\text{max}} \text{tanh}(s \widetilde{\Delta d}),
\end{equation}
where $d_{\text{max}}$ is the maximum value of $\Delta d$ and $s$ is a learnable scale parameter.
The final SDF can be computed as 
\begin{equation}\label{eq:correctedSDF}
    d = -\frac{1}{s^d}\text{logsumexp}\left\{-s^dd_i\right\}_{\{i = 1:P\}} + \Delta d,
\end{equation}
where and $s^d$ is a learnable scaling factor.
Following NeuS~\cite{wang2021neus}, we compute the S-density from the signed-distance function, $d$, and use it with the color estimate of the 3D point to volume render the desired image.
We also regularize the SDF with the Eikonal loss~\cite{gropp2020implicit}:
\begin{equation}\label{eq:eikonal}
    \mathcal L_\text{SDF} = \mathbb{E}[(|\nabla d|_2-1)^2].
\end{equation}

We predict SDFs rather than densities because $\Delta d$ is bounded (Equation~\ref{eq:compress}), and thus it naturally bounds the difference between the estimated surface positions of object $\Omega$ and the ellipsoid approximation, $\ellipsoids$.

\begin{figure}
    \centering
    \vspace*{-5mm}
    \input{figures/reconstruction_figure}
    \vspace*{-5mm}
    \caption{\textbf{Animated figure.} Reconstruction and part segmentation rendered from novel perspectives.}
    \label{fig:reconstruction}
    \vspace{-3mm}
\end{figure}

\subsection{Rendering}\label{sec:rendering}
The color of an output pixel can be predicted by volumetric rendering using the signed-distance function, as proposed in NeuS~\cite{wang2021neus}, and which we briefly describe here for completeness.
The discrete opacity of the $j$-th point along the 3D ray corresponding to the output pixel can be computed as
$\alpha_j = \max \left((\Phi_s(d_j) - \Phi_s(d_{j+1}))/\Phi_s(d_j), 0 \right)$,
where $d_j$ is the signed-distance function at the point (represented in global coordinates) and $\Phi_s$ is a sigmoid function.
From this equation we compute the accumulated transmittance along the ray,
$T_j = \prod_{k=1}^{j-1} (1 - \alpha_k)$,
which we use to estimate the color of the output pixel as $\hat{{\bf C}} = \sum_j T_j \alpha_j \pointcolor_j$.
Similarly, the foreground mask can be rendered as $\hat{\text{M}} = \Sigma_j T_j \alpha_j$.
The photometric reconstruction loss $\mathcal{L}_{\text{photo}}$ is then
\begin{equation}\label{eq:photometric}
    \mathcal{L}_{\text{photo}} = \mathbb{E}_{t, \text{ray}}[||\hat{\bf C} - {\bf C}_{GT}||_2^2 + ||\hat{{\bf M}} - {\bf M}_{GT}||_2^2].
\end{equation}
We refer to the paper by Wang~\etal for more details~\cite{wang2021neus}.

\subsection{Discovery of the 3D Joints}\label{sec:structure}

So far, we have described the object's parts as an \emph{unstructured} set of ellipsoids $\ellipsoids$.
That is, each part's transformation is applied in the global coordinate frame, and the parts act independently of each other.
However, because these ellipsoids represent the object explicitly, they allow us to discover the underlying structure.
Specifically, we make two observations.
First, a point inside part $e_i$ that coincides with (is close to) a point in part $e_j$ as the relative pose between the two parts changes, is likely to be a joint that connects the two parts.
We detail how we leverage this insight in Section~\ref{sec:discovery}.
Because we do not know the number of parts a priori, we start by over-segmenting the object.
Our second insight is that two connected parts that maintain the same relative pose throughout the sequence can be merged:
the joint between them is not necessary to explain the poses observed in the input sequence, Section~\ref{sec:merging}.
While we discover the final structure and finalize the part merging after convergence, we also compute their respective losses at training time for additional regularization (see  Section~\ref{sec:regularization}).

\begin{figure}
    \captionsetup[subfigure]{labelformat=empty}
    \centering
    \includegraphics[width=\columnwidth]{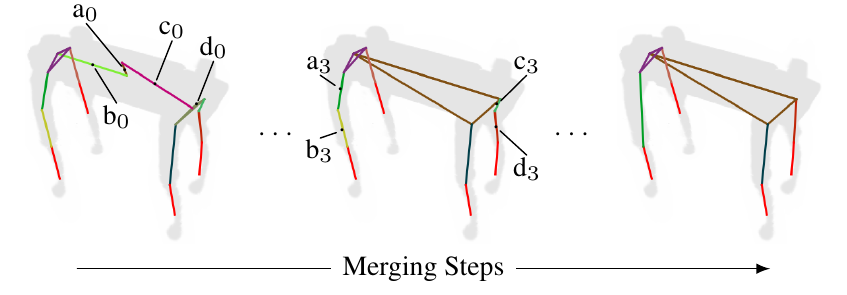}
    \caption{We do not assume the number of parts to be known. Rather, we intentionally over-segment the object for training. After convergence, we merge parts that are static relative to each other. In this case, the first three steps of our procedure merge a$_0$, b$_0$, c$_0$, and d$_0$ into a single part. The final two steps merge a$_3$ with b$_3$, and c$_3$ with d$_3$. A polygon (\eg, the triangular body of the robot) indicates a part that is connected to more than two parts.}
    \label{fig:merging}
\end{figure}

\subsubsection{Structure Discovery}\label{sec:discovery}
We start by sampling N equally spaced joint candidates, ${\joint}_i^n$, for each part $e_i$, see Figure~\ref{fig:parametrization}.
We provide more details about the sampling in the appendix.
In order to discover connections we first compute the distance between all candidates over all frames and for every part pair $(i,j)$
\begin{equation}
    l_{i,j}^{m,n} = \sum_t\left(||\joint_i^m - \joint_j^n||^2_2 + \lambda_l ||\partcenter_i - \partcenter_j||^2_2\right),
\end{equation}
where the second term penalizes connections between parts that are far from each other, $\lambda_l$ is a regularization coefficient, and $t$ is the frame id.
To prevent the distance from changing too quickly, we smooth it across training iterations
\begin{equation}\label{eq:ma_parts_distance}
    \bar{l}^{m,n}_{i,j}(\tau+1) \leftarrow (1-\epsilon)\cdot\bar{l}^{m,n}_{i,j}(\tau) + \epsilon \cdot l^{m,n}_{i,j}(\tau),
\end{equation}
where $\epsilon$ is a momentum, and $\tau$ the training iteration.
We compute the cost of connecting parts $i$ and $j$ as
\begin{equation}
    \bar{l}_{i,j}(\tau) = \min_{n, m} \bar{l}~^{m,n}_{i,j}(\tau).
\end{equation}
We sort the list of $\bar{l}_{i,j}$'s for all parts in ascending order and traverse it to connect the parts that are closest (lowest cost).
We assume the object's structure, $\jointStruct$, to be an acyclic graph, so we require that there be a path between any two joints, and we do not allow connections that would create loops.
We do not connect parts that violate this requirement, even if their $\bar{l}$ is the next lowest.
This procedure allows us to determine the structure of any articulated object that can be modelled as an acyclic graph.
We note that modeling the structure as a tree naturally yields a hierarchy for the parts (with an arbitrary definition of the root node), which is necessary for re-posing.
We also compute the overall cost associated with a particular configuration $\jointStruct$
\begin{equation}\label{eq:structure}
    \mathcal{L}_{\jointStruct} = \sum_{(i, j)\in \jointStruct} \bar{l}_{i,j},
\end{equation}
which we use to regularize our training procedure (see Section~\ref{sec:regularization}).
Figure~\ref{fig:merging} shows the typical
\begin{wrapfigure}[7]{r}[0pt]{0.15\columnwidth}
    \vspace{-5mm}
    \hspace{-6mm}
    \includegraphics[width=0.2\columnwidth]{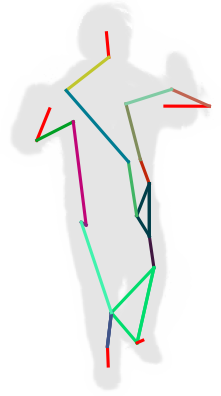}
\end{wrapfigure}
quality of the structure we identify.
Note that a part can connect to multiple parts.
While this approach is reasonably robust, we found that roughly one out of ten random initializations (Section~\ref{sec:trajectory}) can yield to slightly incorrect discovered structures, as shown in the inset on the right.

\subsubsection{Part Merging}\label{sec:merging}
Rather than assuming prior knowledge on the total number of parts, we over-segment the object and merge redundant parts.
Specifically, we combine parts that are static with respect to each other throughout the sequence, see Figure~\ref{fig:merging}.
Differently put, we only preserve the articulations that are necessary to explain a change of pose in the input videos.
The relative position between parts can be computed as $\rotmat_i^j = \rotmat^{-1}_i\rotmat_j$ and $\partcenter_i^j = \rotmat^{-1}_i (\partcenter_j - \partcenter_i)$.
We can then measure the relative motion as
\begin{equation}\label{eq:motion}
    D_{i, j} = \sigma_{t}(\rotmat_i^j) + \lambda_\text{motion}\sigma_{t}(\partcenter_i^j),
\end{equation}
where $\sigma_t$ is the standard deviation over time.
We use Equation~\ref{eq:motion} to define an additional loss term for our training:
\begin{equation}\label{eq:merging}
    \mathcal{L}_{\text{merge}} = \frac{1}{P^2}\sum_{i\neq j}D_{i, j}\Phi_1\left(\frac{\bar{D}-D_{i, j}}{\bar{D}}\right),
\end{equation}
where $\bar{D}$ is a hyperparameter, and $\Phi_1$ is a sigmoid function.
Additional training details can be found in the appendix.
After the training is complete, we merge parts with limited motion relative to each other.
Specifically, we compute Equation~\ref{eq:motion} for all pairs of parts and iteratively merge those for which $D_{i, j}$ is small.
A few steps of this process are shown in Figure~\ref{fig:merging}.

\begin{figure}
    \centering
    \input{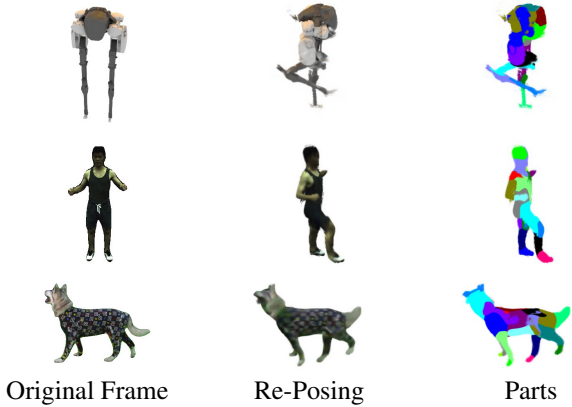}
    \vspace{-5mm}
    \caption{\textbf{Animated figure.} After learning the 3D joints and parts of a previously unseen object category from a multi-view video sequence, we can re-pose it by explicitly manipulating rotation and translation of each joint and part. Credits: human~\cite{peng2021neural}, dog~\cite{rgbd-dogs}.}
    \label{fig:reposing}
    \vspace{-3mm}
\end{figure}

\subsection{Training Strategy and Regularization}\label{sec:regularization}

We train our system on a single scene, and in an end-to-end fashion.
The process optimizes also for the parts' radii $\{{\bf r}_i\}$, in addition to training the parameters of the two MLPs.
To help stabilize the training, we progressively increase the number of frames used for training as the training converges.
This strategy yields a reasonable initialization of the structure, which is then adjusted to capture a consistent part decomposition and structure over the entire video.

Our loss function comprises several terms, including $\mathcal{L}_{\text{SDF}}$, $\mathcal{L}_{\text{photo}}$, $\mathcal{L}_{\jointStruct}$, and $\mathcal{L}_{\text{merge}}$ in Equations~\ref{eq:eikonal}, \ref{eq:photometric}, \ref{eq:structure}, and \ref{eq:merging}, respectively.
However, we only add $\mathcal{L}_{\text{merge}}$ after all the frames are added to the training.
We describe additional regularization terms in the following, and we evaluate the contribution of each term in Section~\ref{sec:ablation}.
The final loss is a weighted sum of these terms, see the appendix.

\paragraph{Ellipsoid Surface Regularization}
Our explicit use of ellipsoids to approximate the shape of the object allows us to sample points from the surface at a low cost.
The projection of sampled surface points onto the image should cover the whole foreground mask of the object, and no pixels outside of it.
We encourage this by minimizing the chamfer distance between the points sampled from the surface and the points sampled from the foreground mask:
\begin{equation}
    \mathcal L_{\mathcal{E}} = \frac{1}{N_\mathcal{E}}\sum_i \min_j ||{\bf p}^\mathcal{E}_i - {\bf p}^M_j||^2_2\nonumber
    + \frac{1}{N_M}\sum_j \min_i ||{\bf p}^\mathcal{E}_i - {\bf p}^M_j||^2_2,
\end{equation}
where ${\bf p}^\mathcal{E}_i$ are the coordinates of points randomly sampled from the surface of $\mathcal{E}$ and projected into the image space, $N_\mathcal{E}$ is their number, ${\bf p}^M_j$ are the coordinates of points randomly sampled from the mask $M_{GT}$, and $N_M$ is their number.

\noindent\textbf{Part Coverage Loss}
Similarly, the centers of the parts, $\partcenter_i$, should be distributed over the foreground mask, rather than being concentrated in a region.
Therefore, the centers of the parts are also learned to minimize the chamfer distance to the foreground mask:
\begin{align}
    \mathcal L_\partcenter = \frac{1}{P}\sum_i \min_j ||\partcenter_i - {\bf p}^M_j||^2_2 + \frac{1}{N_M}\sum_j \min_i ||\partcenter_i - {\bf p}^M_j||^2_2.
\end{align}

\noindent\textbf{Separation Loss}
We further discourage the parts themselves to be concentrated in a single region by penalizing small distances between their centers:
\begin{align}
    \mathcal L_{\text{separation}} = \frac{1}{P^2}\sum_{i\neq j}\exp\left(\frac{|\partcenter_i-\partcenter_j|^2_2}{2\sigma^2}\right),
\end{align}
where $\sigma$ controls the scale of the distances to be regularized.

\section{Evaluation and Results}\label{sec:results}

\begin{figure}
    \centering
    \input{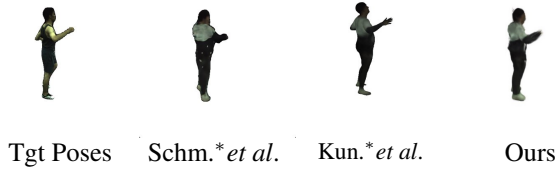}
    \vspace*{-5mm}
    \caption{\textbf{Animated figure.} Re-posing comparison against the baseline methods, Section~\ref{sec:baselines}. For Kundu$^*$~\etal we use the neutral body model.}
    \label{fig:comparisons}
    \vspace{-4mm}
\end{figure}

We evaluate our method's ability to re-pose objects and estimate their joints---both qualitatively and quantitatively.

\subsection{Pose Manipulation}\label{sec:manipulation}
One critical advantage of our explicit representation is the ability to manipulate the pose of previously unseen categories without prior knowledge:
we can directly use the structure we discover by watching the object move.
Given the frame id of a particular sequence, we manipulate the corresponding pose by applying hand-crafted rotations and translations to each of the parts.
To showcase our method's ability to discover the structure, we render a small dataset of seven structurally diverse robots.
We provide more details about the rendered data in the appendix.
We also trained our method on RGBD-dog, dataset of dogs~\cite{rgbd-dogs}.
Figures~\ref{fig:teaser} and~\ref{fig:reposing} show re-posing examples for objects with different structures, number of parts, and joints.
More results are on the project's website.
We do not estimate the range of motion at each joint, and leave it up to the user to define plausible poses.

\subsection{Pose Manipulation for Humans}\label{sec:poseTransfer}
\begin{figure}
    \centering
    \includegraphics[width=1.0\columnwidth]{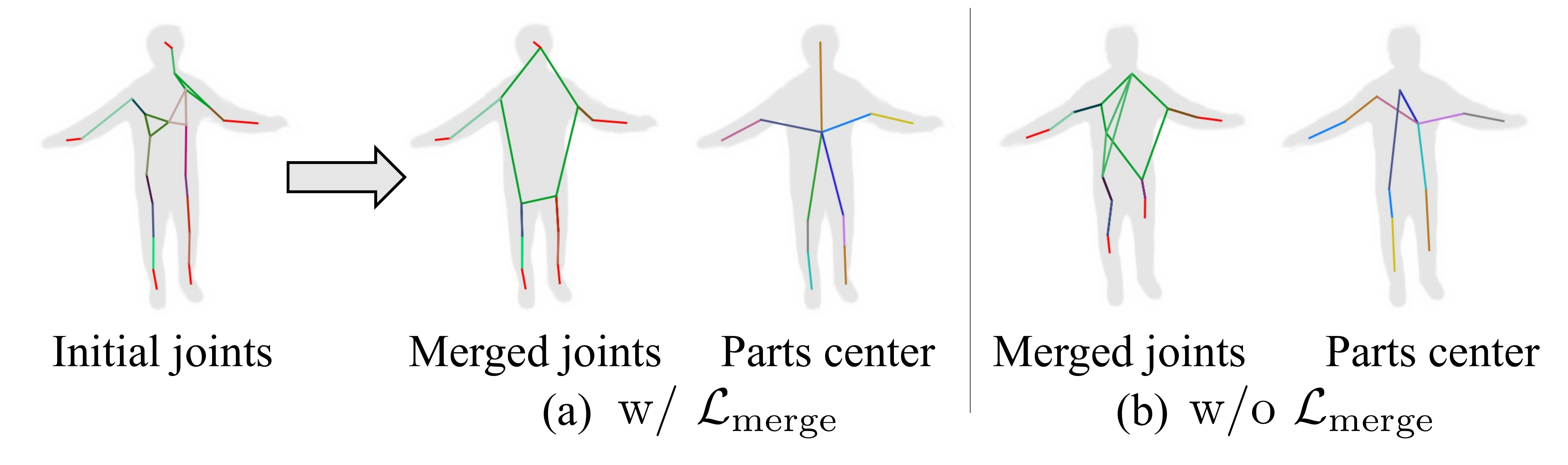}
    \vspace*{-5mm}
    \caption{Effect of merging loss. (a) merging with $\mathcal L_\text{merge}$ and (b) without it.}
    \label{fig:merge_effect}
    \vspace{-7mm}
\end{figure}

The structure our method discovers is plausible, as it allows for accurate re-posing, but it may not coincide exactly with the physical structure of the object.
For a quantitative evaluation we focus on humans, because of the rich literature of methods and  annotated data for this category.
We use the ZJU-MoCap dataset~\cite{peng2021neural} for our experiments.
Specifically, after training our method, we use a subset of the training frames to learn a linear transformation from the joints of an SMPL model~\cite{loper2015smpl} to ours.
Since the ZJU-MoCap dataset~\cite{peng2021neural} has ground truth SMPL annotations, we can use this mapping to re-pose our model to target frames not observed in training, as shown in Figure~\ref{fig:comparisons}.
We use five subjects from the ZJU-MoCap dataset.
For each sequence we use the first $80\%$ of frames for training and the remaining for testing.
The details of the mapping to and from the SMPL model are in the appendix.
Note that this mapping is for evaluation purposes only---our method allows for direct manipulation and does not need a SMPL model, as shown in Figure~\ref{fig:reposing}.
The last two rows of Table~\ref{table:numEval} report the reconstruction quality of our re-posed renderings for the test frames, averaged over all the five subjects.
We provide numbers for our model before and after merging (Section~\ref{sec:merging}).
We note that merging causes a small performance hit because it reduces the expressiveness (DOF) of the representation.
However, even after merging, the performance remains competitive.
Figure~\ref{fig:merge_effect} shows the merging results with and without $\mathcal L_\text{merge}$.
Using $\mathcal L_\text{merge}$ allows us to appropriately pull parts together that have the same motion, and to learn a more meaningful part decomposition.

\newcommand{\best} [1] {\textbf{#1}}
\newcommand{\second} [1] {\underline{#1}}

\begin{table}
    \setlength{\tabcolsep}{1pt}
    \begin{center}
        \caption{Quantitative evaluation.}\label{table:numEval}
        \footnotesize{
            \begin{tabular}{lccccc} 
                \hline
                                                                    & \multicolumn{2}{c}{Novel view} & \multicolumn{2}{c}{Re-posing} & Joints (mm)                                            \\
                                                                    & LPIPS$\downarrow$              & SSIM$\uparrow$                & LPIPS$\downarrow$ & SSIM$\uparrow$ & MPJPE$\downarrow$ \\
                \hline
                Kundu$^*$~\etal\cite{kundu2020appearance}           & 0.059                          & 0.957                         & 0.082             & 0.941          & 12.24             \\
                Schmidtke$^*$~\etal\cite{schmidtke2021unsupervised} & \second{0.055}                 & 0.957                         & \best{0.061}      & \best{0.953}   & 18.56             \\
                \hline
                Ours w/o merge                                      & \best{0.053}                   & \best{0.966}                  & \second{0.064}    & \best{0.953}   & \best{7.59}       \\
                Ours w/ merge                                       & 0.056                          & \second{0.965}                & 0.065             & \second{0.952} & \second{11.11}    \\
                \hline
            \end{tabular}
        }
    \end{center}
    \vspace{-7mm}
\end{table}

\subsection{Baselines Description and Evaluation}\label{sec:baselines}
A direct numerical comparisons with the state-of-the-art is impossible:
ours is the first work that can explicitly re-pose a dynamic object from a previously unseen category, without supervision (other than multi-view supervision), or prior knowledge of the underlying structure.
Moreover existing methods are not scene-specific.
The methods by Schmidtke~\etal~\cite{schmidtke2021unsupervised} and by Kundu~\etal~\cite{kundu2020appearance}, both of which assume a template and only work for humans, are the closest existing solutions for unsupervised, direct pose manipulation.
Although they tackle a more constrained task, we use them as inspiration for baselines that allow for a quantitative evaluation.

Both Schmidtke~\etal~\cite{schmidtke2021unsupervised} and Kundu~\etal~\cite{kundu2020appearance} employ a CNN-based encoder, which allows them to work on scenes not seen in training.
This gives an unfair advantage to our method, which overfits to a specific sequence.
Therefore, we propose the modifications,
which allow us to train both methods for a specific sequence, like ours.

We modify the method by Kundu~\etal by swapping their CNN-based encoder with an MLP that overfits the SMPL parameters to each frame.
These parameters are then used to adapt the SMPL mesh to the pose in the frame.
We train the MLP by enforcing that the color of corresponding vertexes in different frames match.
After convergence we compute the color of all the vertexes of the SMPL model by averaging the colors of the corresponding pixels in all the input frames.
Re-posing their solution, then, reduces to manipulating the SMPL parameters.
To adapt the approach of Schmidtke~\etal, we replace their 2D template with a 3D template and their CNN encoder with an MLP that learns how to deform the 3D template to match the pose at the given time frame.
Given a viewport we can project the template to a 2D representation, which can be converted to an RGB image with a second network.
We denote both baselines with a $^*$ to indicate they are adapted from their original versions, and provide a diagram for each method in the appendix.
A few considerations are in order.
First, the architecture of both MLPs is the same as ours, and the number of the parameters to be predicted comparable.
Second, while we make those methods scene-specific to remove our advantage, they still only work for people and still use a template or an SMPL model, like the original versions.
They are our best effort at a fair comparison.
We train both models on the same train/test split of the same five subjects we use for our method.
For Kundu$^*$~\etal we use the neutral SMPL body model.
A qualitative comparison can be seen in Figure~\ref{fig:comparisons}.
Table~\ref{table:numEval} reports LPIPS~\cite{zhang2018unreasonable} and SSIM~\cite{wang2004image} for both reconstruction (\ie, same pose as in one of the input frames, but different view) and re-posing.
Our method is on par or slightly better than these baselines despite making no assumptions about the structure of the object.

\begin{figure}
    \centering
    \makebox[0mm][s]{\includegraphics[width=0.3\columnwidth]{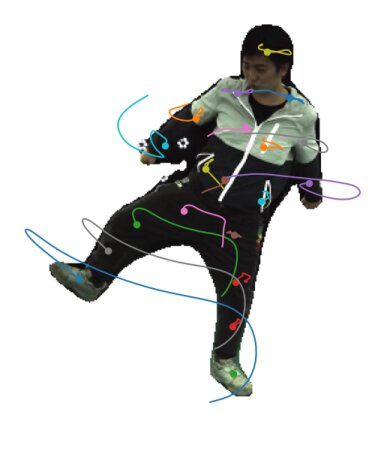}}
    \animategraphics[width=0.3\columnwidth, poster=00024]{10}{figures/tracking/cropped_lowrez_jpg/}{00001}{00028}
    \caption{\textbf{Animated Figure.} The joints our method discovers are plausible and stable across the sequence.}
    \label{fig:tracking}
    \vspace{-3mm}
\end{figure}

\subsection{Joint Estimation Evaluation}\label{sec:jointEval}
Our method discovers plausible joints.
That is, they allow to re-pose the object consistently with the input images, but they may not exactly coincide with the physical joints.
Figure~\ref{fig:tracking} offers a qualitative evaluation:
our 3D joints appear to closely follow the physical joints locations and they are stable over time.
For a quantitative evaluation, we use one tenth of the frames in each sequence to compute a linear mapping from our joints and joints candidates to the joints of the SMPL model provided by the dataset, as is common practice for methods that discover landmarks~\cite{thewlis2017unsupervised, zhang2018unsupervised, lorenz2019unsupervised}.
The details of the algorithm that regresses this mapping are in the appendix.
We apply the linear mapping to the remaining frames to compute the mean per joint position error (MPJPE)~\cite{h36m_pami}.
We also compare with the baselines defined in Section~\ref{sec:baselines}.
For Kundu$^*$~\etal we learn a linear mapping from the predicted SMPL vertices to the GT joints provided by the dataset,
while for Schmidtke$^*$~\etal we compute the same linear mapping as for our method.
Once again, our method performs on par, and sometimes even better despite the additional information available to the baseline methods.


\subsection{Ablation Study}\label{sec:ablation}
In our first ablation study we evaluate the effect of each loss term to the overall performance.
We use subject 366 from the ZJU-MoCap dataset and train our model from scratch by disabling one loss term at the time.
The results are shown in Table~\ref{table:ablation}.
We note that the additional terms have a marginal effect on the quality of the rendered images, but they do reduce the joints estimation error measurably.
Although $L_\partcenter$ slightly degrades the novel view synthesis performance, without it, we observe issues with the structure discovery.
Qualitative results are shown in the appendix.
In our second experiment, we evaluate the importance of training $\decoder$ to predict an SDF residual $\Delta d$, instead of the SDF $d$ itself, as done in Neural-GIF~\cite{tiwari2021neural}.
Table~\ref{table:ablation} confirms that predicting residual is critical to both the image reconstruction quality and the joints estimation.
\begin{table}
    \setlength{\tabcolsep}{3pt}
    \begin{center}
        \caption{Ablation study.}\label{table:ablation}
        \vspace*{-3mm}
        \scalebox{0.75}{
            \begin{tabular}{l|cc|ccccc} 
                \hline
                                             & Ours           & +$\mathcal{L}_{\text{merge}}$ & -$\mathcal{L}_{\Gamma}$ & -$\mathcal{L}_{\mathcal{E}}$ & -$\mathcal{L}_{{\bf t}}$ & -$\mathcal{L}_{\text{separation}}$ & -$\Delta d$ \\
                \hline
                Novel view LPIPS$\downarrow$ & 0.063          & \second{0.062}                & \second{0.062}          & \second{0.062}               & \best{0.061}             & \second{0.062}                     & 0.065       \\
                Novel view SSIM$\uparrow$    & \second{0.958} & \second{0.958}                & \second{0.958}          & \best{0.959}                 & \best{0.959}             & \best{0.959}                       & 0.954       \\
                Novel pose LPIPS$\downarrow$ & \best{0.065}   & \best{0.065}                  & 0.069                   & 0.069                        & \best{0.065}             & \second{0.067}                     & 0.077       \\
                Novel pose SSIM$\uparrow$    & \best{0.954}   & \second{0.953}                & 0.951                   & 0.952                        & \best{0.954}             & 0.952                              & 0.946       \\
                Joint MPJPE(mm)$\downarrow$  & \best{8.49}    & 10.25                         & 9.35                    & 12.13                        & \second{8.70}            & 9.72                               & 22.14       \\
                \hline
            \end{tabular}
        }
    \end{center}
    \vspace{-5mm}
\end{table}

\section{Discussion and Limitations}\label{sec:limitations}
Our method can discover the structure of unseen categories, but it needs a foreground mask, which may not be available for new classes.
While off-the-shelf instance segmentation approaches can be used, as we do for the experiments on humans~\cite{chen2020blendmask} and dogs~\cite{cheng2021maskformer}, this limits the practical applicability of our method.
In addition, our method requires multi-view videos that capture the object from all sides.
The results in the paper use five cameras around the object for robots, six cameras for humans, and eight cameras for the dog.
The four viewpoints available in the Human3.6M dataset~\cite{h36m_pami} do not provide sufficient coverage for our method.
An additional constraint is that we can only re-pose parts that move relative to each other in the training sequence---we cannot infer what we cannot see.
Our solution does not tackle the problem of defining plausible motion ranges around the joints and focuses on spherical joints, leaving different types of joints, such as sliding joints, for future work.
We also observed that the randomness of the structure initialization can sometimes affect the structure discovery (Section~\ref{sec:discovery}). We leave it to future work to find a more elegant solution than simply re-initializing it when this happens.

\noindent\textbf{Societal impact.} Our approach lends itself to similar dishonest uses as deepfakes, \eg, a person could be ``rendered to perform'' illegal, incriminating, or indecent activities.

\section{Conclusions}\label{sec:conclusions}
We presented a method that discovers the structure of an articulated object from arbitrary categories, by watching it move in a multi-view video.
It can then render the object from novel views and even directly manipulate its pose.
Our method works for arbitrary articulated objects, as we show using robots with varying structures.

\section*{Acknowledgments}
This work was partially supported by JST AIP Acceleration Research JPMJCR20U3, Moonshot R\&D Grant Number JPMJPS2011, JSPS KAKENHI Grant Number JP19H01115 and Basic Research Grant (Super AI) of Institute for AI and Beyond of the University of Tokyo. We thank the reviewers for their thoughtful suggestions, which significantly helped us improve our paper.

{\small
\bibliographystyle{ieee_fullname}
\bibliography{biblio}
}

\appendix
\section{Implementation details}
\subsection{Pose Estimation (Section~3.2)}
MLP $\traj$ takes the positionally encoded frame id as input and outputs a vector of $9P$ dimensions.
The $3P$ dimensions correspond to each element of $\partcenter_i$, and the remaining
$6P$ dimensions correspond $\rotmat_i$. For the rotation matrix calculation, please refer to Zhou~\etal~\cite{zhou2019continuity}, Section~B.
$\traj$ is a 4-layer MLP with a hidden dimension of 256.
\subsection{SDF Computation of an Ellipsoid (Section~3.3)}
The merit of using ellipsoids as the representation of the parts is that their SDFs are continuous functions and can be computed cheaply.
In this subsection, we explain how to calculate the SDF of an ellipsoid.
First, the surface of an ellipsoid of radii ${\bf r}$ is given by
\begin{equation}
  f({\bf x}, {\bf r}) = \frac{x^2_1}{r^2_1} + \frac{x^2_2}{r^2_2} + \frac{x^2_3}{r^2_3} = 1,
\end{equation}
where position ${\bf x} = (x_1, x_2, x_3)$ and radii ${\bf r} = (r_1, r_2, r_3)$.
We calculate the SDF of an ellipsoid as follows.
First, from the query point ${\bf x}$, we find the nearest ellipsoid surface point ${\bf x}_{e}$.
Since this cannot be solved analytically, we use Lagrange's multiplier method and Newton's method to find the point. The cost function $\mathcal{L}$ is defined as
\begin{equation}
  \mathcal{L} = |{\bf x} - {\bf x}_{e}|^2_2 - \lambda(f({\bf x}_e, {\bf r}) - 1),
\end{equation}
and we solve $\frac{\partial \mathcal{L}}{\partial {\bf x}_e} = \frac{\partial \mathcal{L}}{\partial \lambda} = 0$. This can be transformed into the following equations
\begin{numcases}
  {}
  {\bf x}_{e} = {\bf r}^2 \oslash ({\bf r}^2 + \lambda) \odot {\bf x}\label{eq:ellipsoid_surface}\\
  |{\bf r} \oslash ({\bf r}^2 + \lambda) \odot {\bf x}|^2_2 = 1,\label{eq:newton}
\end{numcases}
where $\oslash$ is element-wise division, $\odot$ is element-wise product, and ${\bf a}^2 = {\bf a}\odot {\bf a}$.
We use Newton's method to find the largest solution for $\lambda$ in Equation~\ref{eq:newton} and substitute it into Equation~\ref{eq:ellipsoid_surface} to get ${\bf x}_{e}$.

The distance between the searched point and the input point ${\bf x}$ is the absolute value of the SDF, which has a negative sign when ${\bf x}$ is inside the ellipsoid and a positive sign when it is outside:
\begin{equation}
  \text{SDF}({\bf x}, {\bf r}) = \text{sign}(f({\bf x}, {\bf r}) - 1) |{\bf x} - {\bf x}_{e}|_2.
\end{equation}
However, since ${\bf x}_{e}$ is computed numerically, the gradient is not propagated to ${\bf r}$.
Therefore, we re-parametrize ${\bf x}_{e}$ using ${\bf r}$ by projecting ${\bf x}_{e}$ onto the surface of the unit sphere and back onto the surface of the ellipsoid.
\begin{equation}
  \tilde{{\bf x}}_{e} = ({\bf x}_{e}\oslash{\bf r}).\text{detach}()\odot {\bf r},
\end{equation}
where $\oslash$ is element-wise division, $\odot$ is element-wise product, and detach() is a stop-gradient operation.

Finally, the differentiable SDF of an  ellipsoid is computed as,
\begin{equation}
  \text{SDF}({\bf x}, {\bf r}) = \text{sign}(f({\bf x}, {\bf r}) - 1) |{\bf x} - \tilde{{\bf x}}_{e}|_2.
\end{equation}

\subsection{Shape and Appearance Decoder (Section~3.3)}
MLP $\decoder$ takes a feature vector ${\bf f}$ at a query 3D location ${\bf x}^g$ and outputs the color ${\bf c}$ and a residual SDFs $\widetilde{\Delta d}$.
The color on the surface of real objects changes in complex ways according to the time, pose,  and view direction.
We simplify the problem by assuming the color depends only on ${\bf f}$, meaning the color is constant across view and time.
$\decoder$ is a 8-layer MLP with a hidden dimension of 256.

\subsection{Joint Candidates (Section~3.5.1)}
In this subsection, we explain the details of the joint candidates defined in the Section~3.5.1 in the main paper.
When a point inside one ellipsoid $e_i$ is close to a point inside another ellipsoid $e_j$ throughout the entire video, the point is considered to be a joint between parts $i$ and $j$.
To find these points, we create several joint candidate points $\{\joint^n_i\}_{\{n=1:N\}}$ inside the ellipsoids in advance, and minimize the distance between them.
We define six candidates inside each ellipsoid in the local coordinate, as follows
\begin{equation}
  \hat{\joint}^n_i = {\bf r}_i \odot {\bf \joint}^n,
\end{equation}
where ${\bf \joint}^n\in \{(\pm \frac{3}{4}, 0, 0), (0, \pm \frac{3}{4}, 0), (0, 0, \pm \frac{3}{4})\}$.
This is then transformed into the global coordinate system using the rotation and translation of each part
\begin{equation}
  \joint^n_i = \rotmat_i \hat{\joint}^n_i + \partcenter_i.
\end{equation}

\subsection{Frame scheduling (Section~3.6)}\label{sec:frame_scheduling}
In order to stabilize the training, we progressively increase the number of frames used for training. First, we use the first $T_0$ frames of data to train for $\tau_0$ iterations. Then, the number of frames used for training is increased linearly so that all frames $T$ of the video are used at $\tau_1$ iteration. After that, all frames are used for training until $\tau_\text{final}$. In the experiment using human data, we set $T_0=10$, $\tau_0=10\text{k}$ $\tau_1=80\text{k}$.

\subsection{Loss (Section~3.6)}
The loss function is a weighted sum of $\mathcal{L}_{\text{SDF}}$, $\mathcal{L}_{\text{photo}}$, $\mathcal{L}_{\Gamma}$, $\mathcal{L}_{\text{merge}}$, $\mathcal{L}_{\mathcal{E}}$, $\mathcal{L}_\partcenter$, and $\mathcal{L}_{\text{separation}}$:
\begin{align}
  \mathcal{L}_{\text{total}} = & \lambda_{\text{SDF}}\mathcal{L}_{\text{SDF}} + \lambda_{\text{photo}}\mathcal{L}_{\text{photo}} + \lambda_{\Gamma}\mathcal{L}_{\Gamma} +                                                                       \\\notag
                               & \lambda_{\text{merge}}\mathcal{L}_{\text{merge}} + \lambda_{\mathcal{E}}\mathcal{L}_{\mathcal{E}} + \lambda_{\partcenter}\mathcal{L}_\partcenter + \lambda_{\text{separation}}\mathcal{L}_{\text{separation}}.
\end{align}
We used $\lambda_{\text{SDF}}=0.2$, $\lambda_{\text{photo}}=1$, $\lambda_{\text{merge}}=0$, $\lambda_{\mathcal{E}}=600$, $\lambda_{\partcenter}=1000$, and $\lambda_{\text{separation}}=1$.
We gradually increase $\mathcal{L}_{\Gamma}$ from $2$ to $50$ until iteration $\tau_0$ for training stability.
From iteration $\tau_2$, we set $\lambda_{\text{merge}}=5$ and train the model until $\tau_{\text{final}}$. We set $\tau_2=150\text{k}$.
\subsection{Other Training Details}
The other hyper-parameters were set as
$d_{\text{max}}=0.02$, $\lambda_l=0.02$, $\lambda_\text{motion}=3$, $\bar{D}=0.1$, $\epsilon=0.01$. For the weighted positional encoding in Equation~4 in the main paper, we apply positional encoding~\cite{mildenhall2020nerf} to spatial locations ${\bf x}_i$ with 6 frequencies following the training setting of NeuS~\cite{wang2021neus}.

We used AdamW optimizer~\cite{loshchilov2018decoupled} with learning rate 0.0003, and $\beta_1=0.9$, $\beta_2=0.999$, $\lambda = 0.005$. All training images are resized to $512\times512$.
We train the model up to $\tau_{\text{final}}=200\text{k}$ iterations with a batch size of 16. We randomly sample 384 rays from each image for human data, where $P$ is set to 20. Training takes about 48 hours on a single NVIDIA A100 GPU.

\subsection{Pose Manipulation (Section~4.1)}
Since our method estimates explicit joint relationships between parts, we can freely manipulate the pose of the object. First, the position of the final joints $\joint_{i, j}$ is defined as the midpoint of the candidate points $\joint_i^m$ and $\joint_j^n$ connected in Equations~16 and 17 of the main paper:
\begin{equation}
  \joint_{i, j} = \frac{1}{2}(\joint_i^m + \joint_j^n)\quad \text{s.t.}\;(m,n) = \text{arg}\min_{m, n} \bar{l}~^{m,n}_{i,j}.
\end{equation}
By manually rotating the part $i$ or $j$ and its children parts around this joint $\joint_{i, j}$, the pose of the object can be freely changed, and novel rotations and translations for each part can be obtained $\{\rotmat_i, \partcenter_i\}^{\text{novel }}_{\{i=1:P\}}$. To render the novel pose image, we directly input $\{\rotmat_i, \partcenter_i\}^{\text{novel }}_{\{i=1:P\}}$ to the second network $\decoder$.

\subsection{Baselines (Section~4.3)}

\begin{figure*}
  \centering
  \subfloat[Kundu$^*$\etal~\cite{kundu2020appearance}]{\includegraphics[width=.99\columnwidth,valign=c]{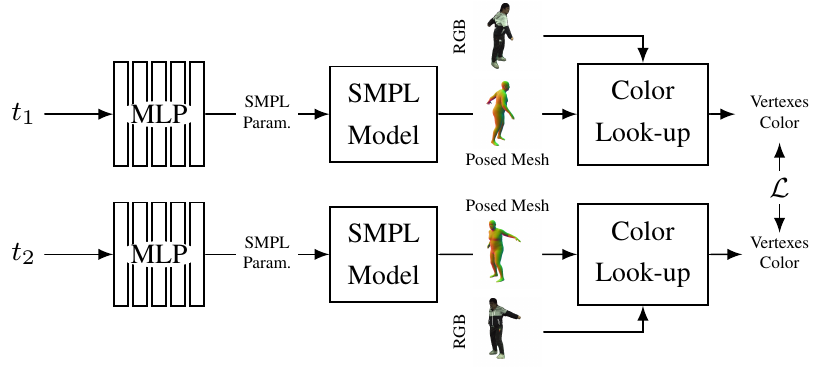}}\hspace{10pt}
  \subfloat[Schmidtke$^*$\etal~\cite{schmidtke2021unsupervised}]{\includegraphics[width=.99\columnwidth,valign=c]{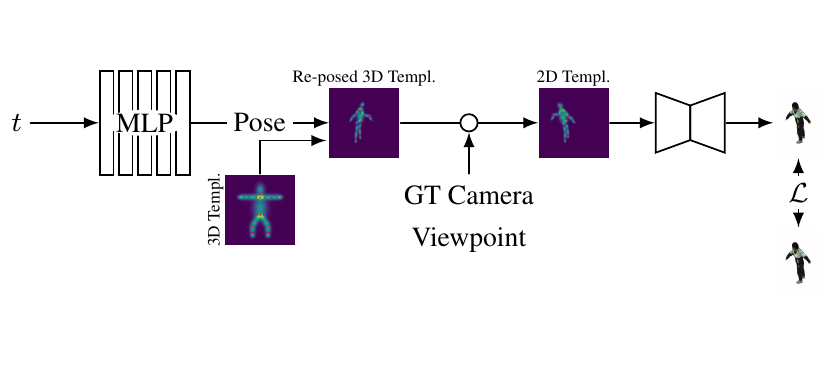}}
  \caption{Baselines used for comparison. The $^*$ indicates that these are adaptations of the original methods.}
  \label{fig:baselines}
  \vspace{-3mm}
\end{figure*}

\paragraph{Kundu*~\etal~\cite{kundu2020appearance}}
Their original model estimates SMPL parameters in an unsupervised manner as follows.
First, a CNN-based encoder receives an image and estimates the parameters of the SMPL model.
Based on the estimated parameters, the SMPL mesh is deformed and a pixel value of the image is assigned to each vertex according to its position in the image.
The model is trained to match the colors of the estimated mesh vertices from different images of the same person at different times.
By using CNN-based encoder, they can reconstruct the mesh from unseen monocular images.
However, our method uses videos of specific scenes from multiple viewpoints, which gives us an unfair advantage.
In order to allow for a fair comparison between our method and theirs, we replaced their CNN-based encoder with an MLP $\traj^{\text{Kundu*}}$ that takes frame id as input and estimates SMPL parameters for each frame.
\begin{equation}
  \traj^{\text{Kundu*}}: \gamma(t)\rightarrow {\bf R}(t), {\bf t}(t), \boldsymbol \theta (t), \boldsymbol \beta,
\end{equation}
where ${\bf R}(t)$ is a global orientation, ${\bf t}(t)$ is a global translation, $\boldsymbol \theta (t)$ is joint poses, and $\boldsymbol \beta$ is a shape parameters.
An overview of the method is visualized in Figure~\ref{fig:baselines} (a).
This MLP allows Kundu$^*$~\etal to overfit to the specific video sequence, like our method does.
Please note that $\boldsymbol \beta$ is not time dependent.
The structure of the MLP $\traj^{\text{Kundu*}}$ is the same as that of $\traj$ used in the proposed method except for the output dimension.
Since the authors do not publish their training implementation, all modules and loss functions are our replicated implementation.
For the human pose prior, instead of training the adversarial auto-encoder, we used the pre-trained human pose prior from Bogo~\etal~\cite{bogo2016keep}.
Also, since our experimental setup uses multi-view videos and overfits to a single subject, we do not use reflectional symmetry or shape-consistency loss. Please refer to Kundu~\etal~\cite{kundu2020appearance} for more details.
In addition, since human foreground masks are available in our experimental setup, we use a differentiable renderer~\cite{ravi2020pytorch3d} to render the mask of the mesh and train it so that the L2 norm with the GT foreground mask is small.
We apply the same frame scheduling as in Section~\ref{sec:frame_scheduling} for training stabilization.

After training the model, we assign colors to the vertices of the SMPL mesh for novel view and pose synthesis, where
we average the estimated vertex color for various frame ids and viewpoints using the learned SMPL poses.
\paragraph{Schmidtke*~\etal~\cite{schmidtke2021unsupervised}}
Their original model trains the deformation of a 2D template of a person's structure in an unsupervised manner using image reconstruction.
They use a CNN-based encoder to estimate the 2D deformation parameters.
To extend the method to 3D, we replaced the 2D templates with 3D templates, where the shape is approximated with 3D gaussians.
For a fair comparison, we also replaced the CNN-based encoder with an MLP $\traj^{\text{Schmidtke*}}$ that takes a frame id as input and outputs the deformation parameters of the template in the global coordinate, as in Kundu*~\etal.
\begin{equation}
  \traj^{\text{Schmidtke*}}: \gamma(t)\rightarrow \{\rotmat_i(t), \partcenter_i(t),{\bf s}_i\}_{\{i=1:18\}},
\end{equation}
where $\rotmat_i(t)$ and $\partcenter_i(t)$ are rotation and translation for each part, and ${\bf s}_i$ is 3D scale parameters for each part.
An overview of the method is visualized in Figure~\ref{fig:baselines} (b).
We replace the affine transformation $\boldsymbol \Theta_i$ defined in Equation~3 in Schmidtke~\etal~\cite{schmidtke2021unsupervised} with a physically meaningful transformation
\begin{equation}
  \boldsymbol \Theta =
  \begin{bmatrix}
    s_1R_{1,1} & s_1R_{1,2} & s_1R_{1,3} & t_1 \\
    s_2R_{2,1} & s_2R_{2,2} & s_2R_{2,3} & t_2 \\
    s_3R_{3,1} & s_3R_{3,2} & s_3R_{3,3} & t_3 \\
    0          & 0          & 0          & 1   \\
  \end{bmatrix}.
\end{equation}
Please note that ${\bf s}_i$ is not time dependent.
The deformed 3D template is projected onto a 2D heatmap using the viewpoint of the training image, and then transformed into an RGB image using a second CNN-based network.
Since the model is overfitted on a single scene, we do not input the reference frame to the second network, but only the 2D heatmap.
We modified the implementation based on their public training code\footnote{\url{https://github.com/lschmidtke/shape_templates}}, and the loss function used for training is exactly the same as the original.
Please refer to Schmidtke~\etal~\cite{schmidtke2021unsupervised} for more details.

We apply the same frame scheduling as in Section~\ref{sec:frame_scheduling} for training stabilization.
\subsection{Regression of GT SMPL Pose (Section~4.2, 4.4)}
For human re-posing we learn a linear mapping from the GT SMPL poses to estimated joints and for joint evaluation we learn a mapping from the estimated joints to the GT SMPL poses.
\paragraph{Mapping from SMPL}
For re-posing, we perform a regression from the GT SMPL mesh to our part pose $\rotmat_i$ and $\partcenter_i$.
First, for all frames of the training data, we optimize the linear transformation ${\bf X}$ from the mesh vertices ${\bf V}(t)$ to the concatenation of the learned part centers and  the candidate points ${\bf P}(t) = \text{CAT}\{\partcenter_i, \joint^1_i, \ldots, \joint^N_i\}_{\{i=1:P\}}(t)$ using the least-squares method
\begin{equation}
  \min_{\bf X} \left(\sum_t |{\bf X}{\bf V}(t) - {\bf P}(t)|_F + \frac{1}{2}\lambda |{\bf X}|_F\right).
\end{equation}
When re-posing with SMPL meshes, we compute the part centers and candidate points ${\bf P}^{\text{new}} = \text{CAT}\{\partcenter_i, \joint^1_i, \ldots, \joint^N_i\}^{\text{new}}_{\{i=1:P\}}$ corresponding to the novel pose SMPL mesh ${\bf V}^{\text{new}}$ using the learned linear transformation ${\bf X}$,
\begin{equation}
  {\bf P}^{\text{new}} = {\bf X}{\bf V}^{\text{new}}.
\end{equation}
The part centers $\partcenter_i^{\text{new}}$ of the new pose are obtained by extracting the corresponding elements of ${\bf P}^{\text{new}}_i$.
The rotation matrix $\rotmat^{\text{new}}_i$ is obtained by solving the following optimization:
\begin{equation}
  \min_{\rotmat^{\text{new}}_i} |\rotmat^{\text{new}}_i \hat{\Xi}_i - \Xi_i|_F\quad s.t.\;(\rotmat^{\text{new}}_i)^\top \rotmat^{\text{new}}_i = {\bf I},
\end{equation}
where $\hat{\Xi}_i = \left[\hat{\joint}^1_i, \ldots, \hat{\joint}^N_i\right]$, $\Xi_i = \left[\joint^1_i - \partcenter_i, \ldots, \joint^N_i - \partcenter_i\right]$.
By using the resulting $\partcenter_i^{\text{new}}$ and $\rotmat^{\text{new}}_i$ into the second network $\decoder$ directly, we can re-pose the object.

Similarly, for the re-posing of the baseline Schmidtke*~\etal~\cite{schmidtke2021unsupervised}, the part center $\partcenter_i$ of the 3D template and the points around it
\begin{equation}\label{eq:pseudo_cand}
  \joint^n_i\in\{\partcenter_i\pm r \rotmat_i^1, \partcenter_i\pm r\rotmat_i^2, \partcenter_i\pm r\rotmat_i^3\}
\end{equation}
are used to learn the same linear mapping, where $\rotmat_i = [\rotmat_i^1,\rotmat_i^2,\rotmat_i^3]$ and $r=0.1$.

\paragraph{Mapping to SMPL}
To evaluate the joint, we regress the translation ${\bf J}(t) = \text{CAT}\{{\bf j}_i\}_{\{i=1:23\}}(t)$ of the GT SMPL joints at frame $t$ from the learned object poses at $t$, where CAT is the concatenation operator.
We obtain a linear mapping ${\bf X}$ from the learned poses to the SMPL joints by solving the following optimization problem:
\begin{equation}
  \min_{{\bf X}} \left(\sum_{t\in T_{\text{train}}}|{\bf X}{\bf P}(t) - {\bf J}(t)|_F + \frac{1}{2}\lambda |{\bf X}|_F\right),
\end{equation}
where $T_{\text{train}}$ is the set of frames used for this optimization,
which are uniform sample of $10\%$ of the available frames.

For joint evaluation, a learned linear transformation ${\bf X}$ was applied to the remaining frames to compute the mean per joint position error (MPJPE)~\cite{h36m_pami} between the regressed joint position and the GT joint position.

Similarly, for the baseline Schmidtke*~\etal~\cite{schmidtke2021unsupervised}, we learn a linear mapping from the part center $\partcenter_i$ of the 3D template and the points around it $\joint_i^n$ defined in Equation~\ref{eq:pseudo_cand} to the GT SMPL joints.

For the evaluation of Kundu*~\etal~\cite{kundu2020appearance}, we learn a linear regression ${\bf X}$ from the learned SMPL mesh vertices ${\bf V}_{\text{Kundu}}$ to the GT SMPL joints
\begin{equation}
  \min_{{\bf X}} \left(\sum_{t\in T_{\text{train}}}|{\bf X}{\bf V}_{\text{Kundu}}(t) - {\bf J}(t)|_F + \frac{1}{2}\lambda |{\bf X}|_F\right).
\end{equation}
We evaluated both models in the same way as ours.

\begin{figure}[t]
  \includegraphics[width=\linewidth]{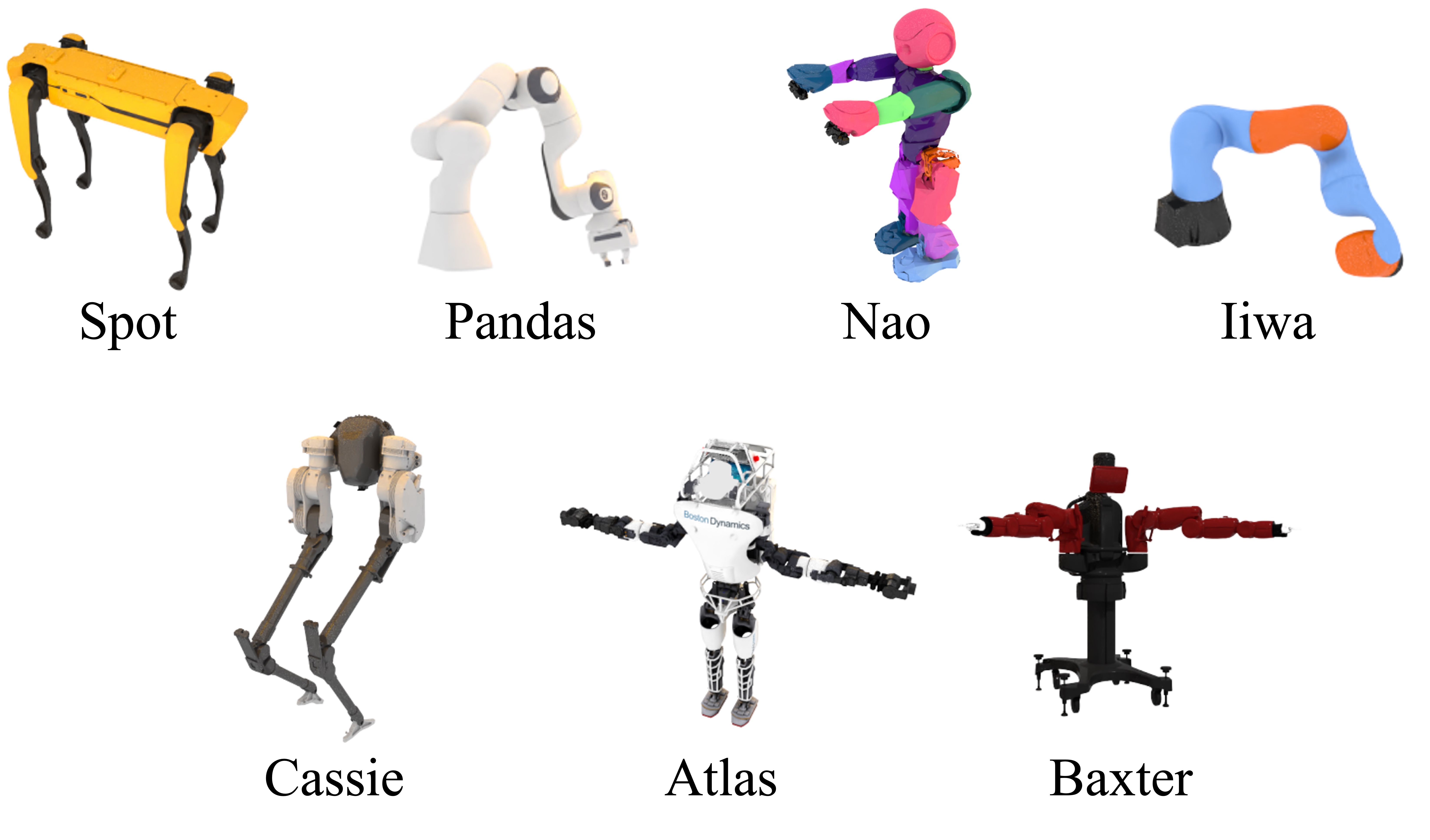}
  \caption{The robots in our datasets.}
  \label{fig:robot}
\end{figure}

\subsection{Effect of \texorpdfstring{$\mathcal L_\partcenter$}{TEXT} (Section 4.5)}
\begin{figure}[t]
  \centering
  \includegraphics[width=0.8\linewidth]{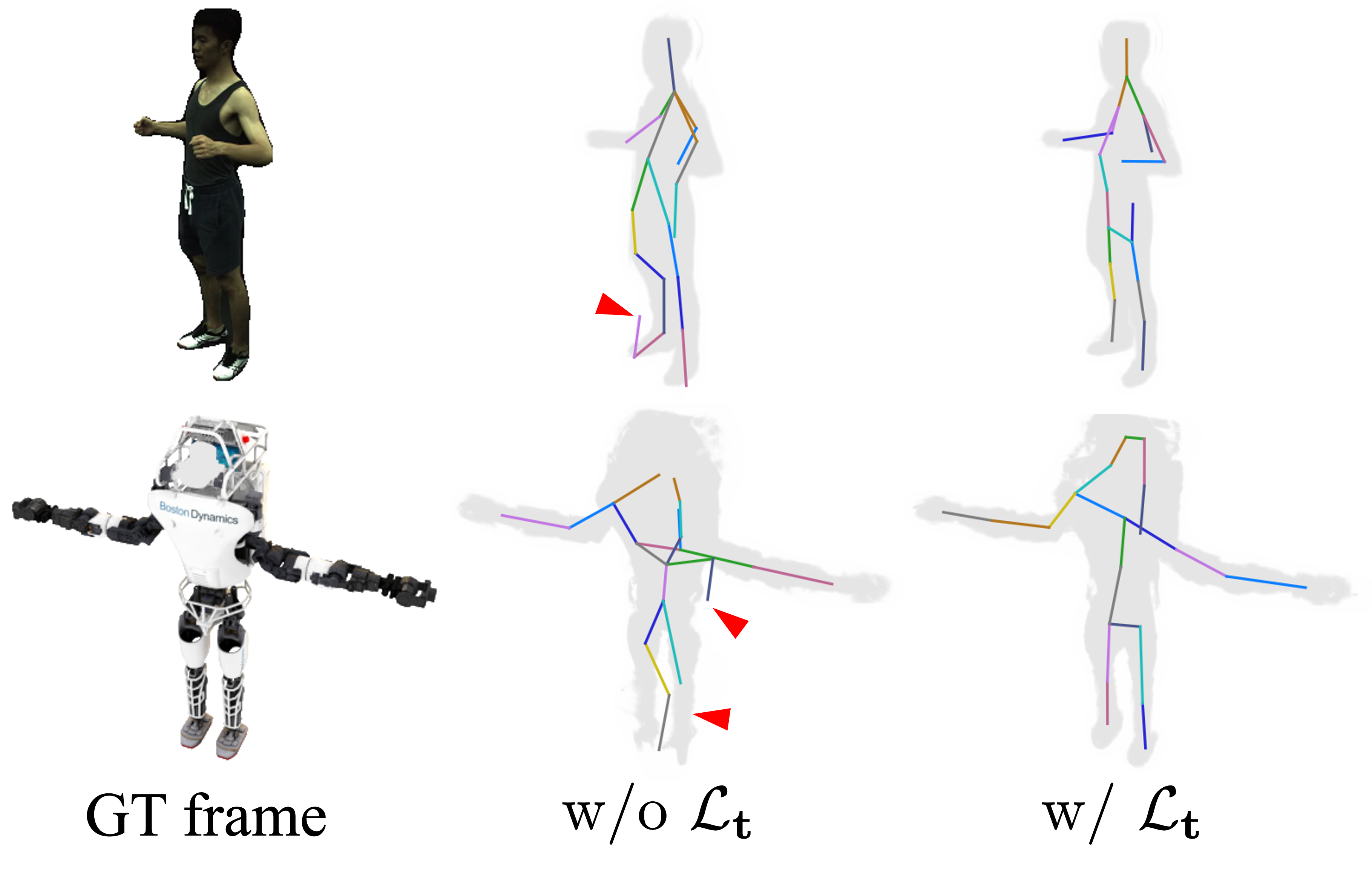}
  \caption{Learned structures with and without $\mathcal L_\partcenter$. Each line connects the center of each part. Red marks show the parts that should be noted.}
  \label{fig:l_partcenter}
\end{figure}

Qualitative results with and without $\mathcal L_\partcenter$ are shown in Figure~\ref{fig:l_partcenter}. We can see that the parts sometimes do not cover the entire object or go outside of the object without $\mathcal L_\partcenter$. This result demonstrates the effectiveness of $\mathcal L_\partcenter$.

\section{Training on the RGBD-Dog Dataset}
We used RGBD-Dog dataset~\cite{rgbd-dogs} for training the dog model. We apply off-the-shelf instance segmentation model~\cite{cheng2021maskformer} to obtain the ground truth masks. In this dataset, it is not possible to isolate the exact region of the dog due to occlusion by a person or overlapping background objects. Therefore, we modified the loss function to ignore occluded regions and regions where the estimated mask is unreliable. Specifically, when calculating $\mathcal L_\text{photo}$, $\mathcal L_\ellipsoids$, and $\mathcal L_\partcenter$, we replace the loss values with 0 for elements that use unreliable pixels in their calculations.

\section{Robot dataset}
In order to demonstrate the applicability of our method to objects with various structures, we created a dataset of robots with seven different structures, see Figure~\ref{fig:robot}.
The dataset consists of 1000 frames of synchronized video with 20 viewpoints per robot.
We sampled five of these views and trained on the first 300 frames of each video.

In order to generate the dataset we use a recent python-based renderer, NViSII~\cite{morrical2021nvisii}.
We use robots that are freely available and have URDF associated with 3D meshes.
In order to animate the robot, we use PyBullet~\cite{coumans2019:pybullet}.
The robot is given random joint goals, and once it reaches these goals,
we repeat the process of giving it random joint goals.
We place 20 fixed cameras on the hemisphere at a fixed distance from the robot, and
we add a warm sunlight to add more light to the scene.
Each frame is rendered with 2000 samples per pixel at 512$\times$512 resolution.
We use the OptiX denoiser to clean the final renders to provide noise free images.
Both the datasets and the scripts to generate the multiview animated objects
will be available at \url{https://github.com/NVlabs/watch-it-move}.

\section{Additional Results for Parts Merging}
\begin{figure*}[t]
  \center
  \includegraphics[width=0.8\linewidth]{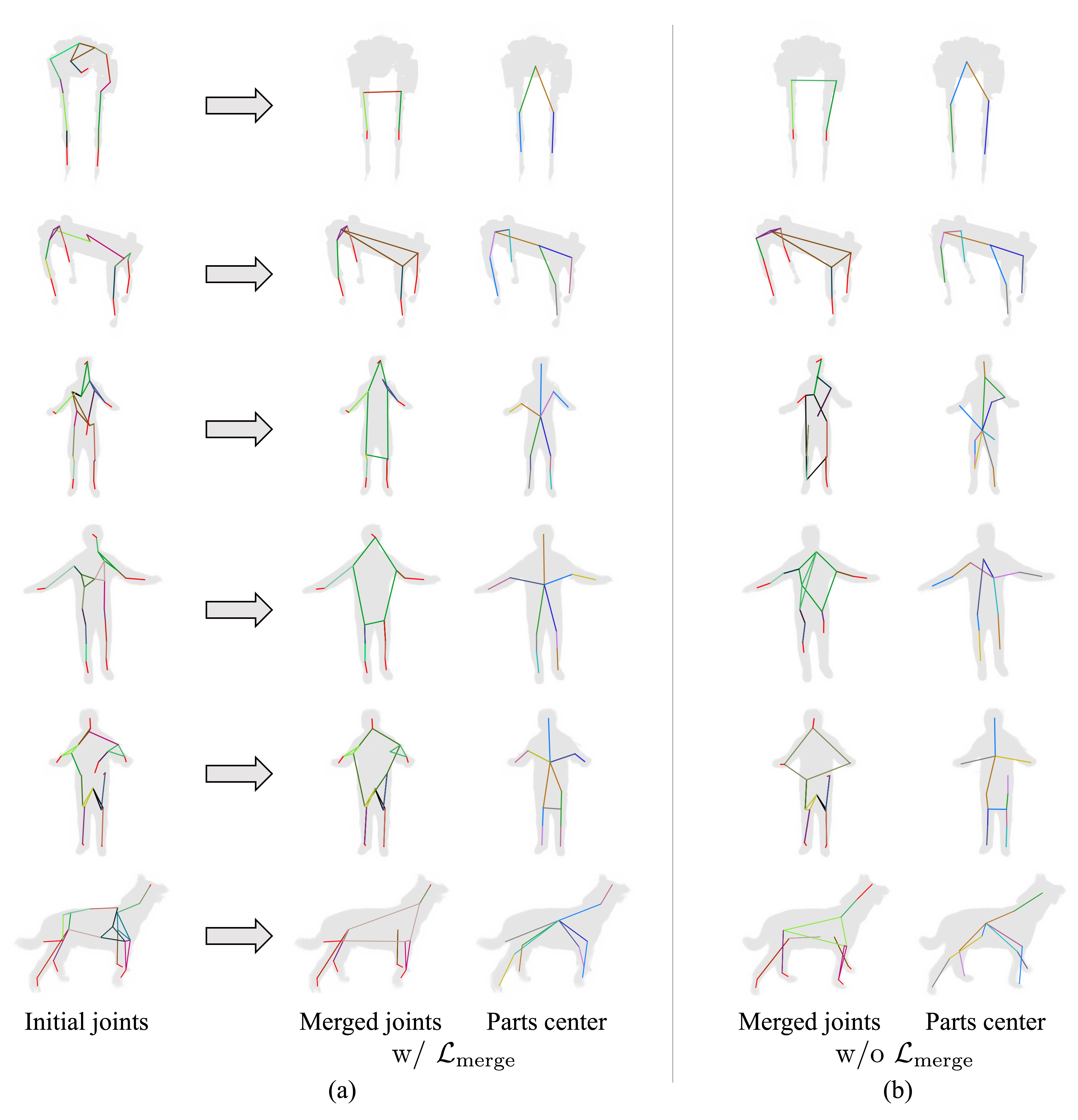}
  \caption{(a) Additional results for part merging. From left to right, we show the initial joints and structure, the joints and structure after merging, and the connection between the centers of the parts. Since there is no joint at the endpoint, the center of the part is connected instead (red lines). A polygonal path indicates a part that is connected to multiple parts. The centers of the parts are shown for clarity. (b) Joints and part centers obtained when $\mathcal{L}_{\text{merge}}$ is disabled.}
  \label{fig:merge_example}
\end{figure*}

Additional results of parts merging are shown in Figure~\ref{fig:merge_example} (a). The results confirm that meaningful parts and the structure are obtained by merging.
To show the effect of $\mathcal{L}_{\text{merge}}$, we show the merging results when it is disabled in Figure~\ref{fig:merge_example} (b).
It can be seen that by using $\mathcal{L}_{\text{merge}}$, we can appropriately pull parts together that have the same relative motion, and learn more meaningful decomposition of parts.

\end{document}